\documentclass[10pt,twocolumn,letterpaper]{article}

\usepackage[scaled]{helvet}
\usepackage[T1]{fontenc}

\usepackage{amsmath}
\usepackage{amssymb}
\usepackage{cvpr}
\usepackage{epsfig}
\usepackage{graphicx}
\usepackage{stackengine}
\usepackage{times}

\usepackage[pagebackref=true,breaklinks=true,colorlinks,bookmarks=false,linkcolor=black,citecolor=black]{hyperref}

\cvprfinalcopy 

\def\cvprPaperID{2343} 

\DeclareMathOperator{\argmax}{argmax}
\DeclareMathOperator{\argmin}{argmin}
\DeclareMathOperator{\far}{FAR}
\DeclareMathOperator{\val}{VAL}
\DeclareMathOperator{\fa}{FA}
\DeclareMathOperator{\ta}{TA}

\newcommand\email[1]{\small{\href{mailto:#1}{\color{black}{\nolinkurl{#1}}}}}

\ifcvprfinal\fi
\begin{document}

\title{FaceNet: A Unified Embedding for Face Recognition and Clustering}

\author{Florian Schroff\\
\email{fschroff@google.com}\\
Google Inc.\\
\and
Dmitry Kalenichenko\\
\email{dkalenichenko@google.com}\\
Google Inc.\\
\and
James Philbin\\
\email{jphilbin@google.com}\\
Google Inc.\\
}
\maketitle

\begin{abstract}
\vspace{-1em}
  Despite significant recent advances in the field of face
  recognition~\cite{gaussianface,deepid2,deepid2plus,deepface}, implementing
  face verification and recognition efficiently at scale presents serious
  challenges to current approaches. In this paper we present a system, called
  FaceNet, that directly learns a mapping from face images to a compact
  Euclidean space where distances directly correspond to a measure of face
  similarity. Once this space has been produced, tasks such as face
  recognition, verification and clustering can be easily implemented using
  standard techniques with FaceNet embeddings as feature vectors.

  Our method uses a deep convolutional network trained to directly optimize the
  embedding itself, rather than an intermediate bottleneck layer as in previous
  deep learning approaches. To train, we use triplets of roughly aligned
  matching / non-matching face patches generated using a novel online triplet
  mining method. The benefit of our approach is much greater representational
  efficiency: we achieve state-of-the-art face recognition performance using
  only \mbox{128-bytes} per face.

  On the widely used Labeled Faces in the Wild (LFW) dataset, our system
  achieves a new record accuracy of \textbf{99.63\%}. On YouTube Faces DB it
  achieves \textbf{95.12\%}. Our system cuts the error rate in comparison to
  the best published result~\cite{deepid2plus} by 30\% on both datasets.

  We also introduce the concept of \emph{harmonic} embeddings, and a
  \emph{harmonic} triplet loss, which describe different versions of face
  embeddings (produced by different networks) that are compatible to each other
  and allow for direct comparison between each other.
\end{abstract}

\vspace{-1em}
\section{Introduction}

\begin{figure}[t]
\begin{center}
\renewcommand\tabcolsep{1pt}
\begin{tabular}{ccc}
 \raisebox{-.5\height}{\includegraphics[width=.4\linewidth]{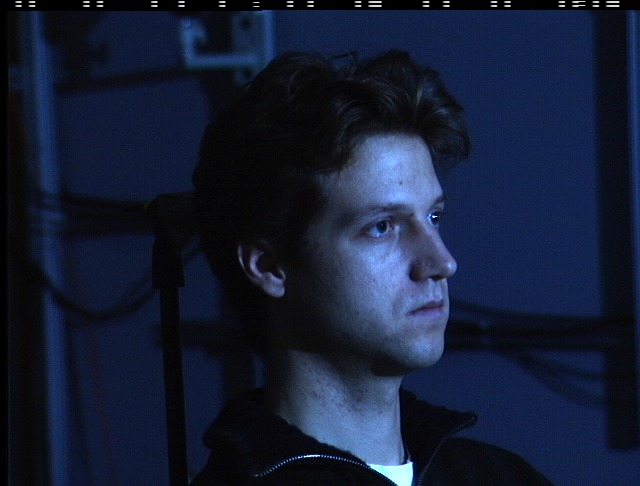}} &
 1.04 &
 \raisebox{-.5\height}{\includegraphics[width=.4\linewidth]{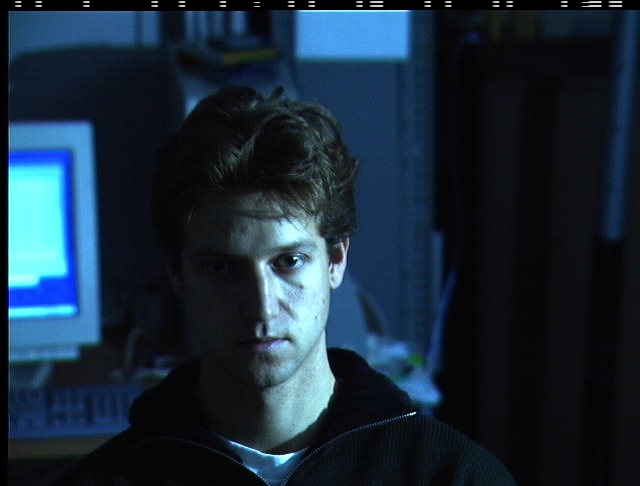}} \\
 1.22 & & 1.33 \\
 \raisebox{-.5\height}{\includegraphics[width=.4\linewidth]{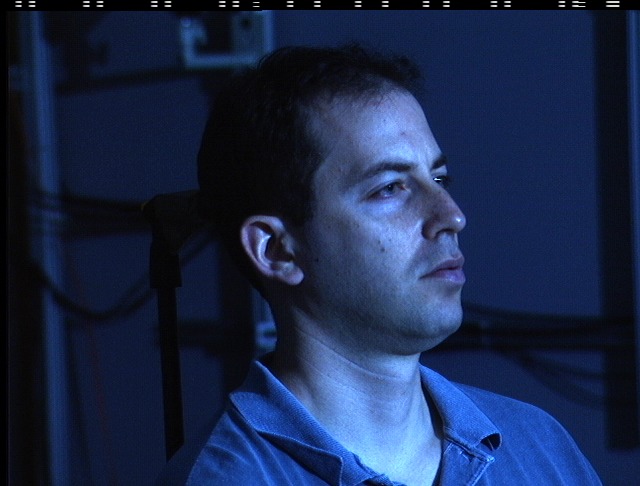}} &
 0.78 &
 \raisebox{-.5\height}{\includegraphics[width=.4\linewidth]{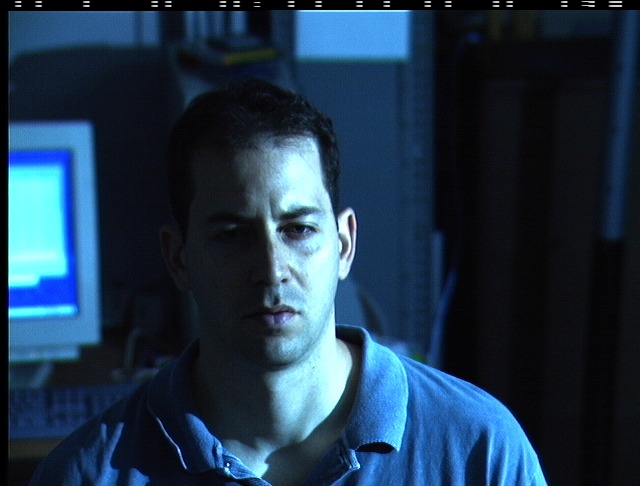}} \\
 1.33 & & 1.26 \\
 \raisebox{-.5\height}{\includegraphics[width=.4\linewidth]{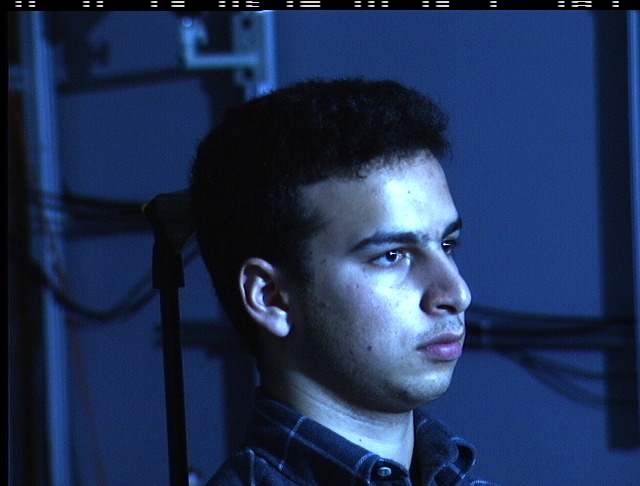}} &
 0.99 &
 \raisebox{-.5\height}{\includegraphics[width=.4\linewidth]{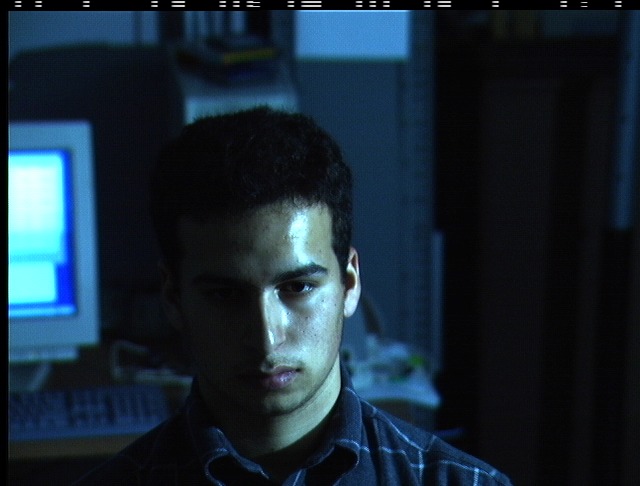}} \\
\end{tabular}
\end{center}
\caption{{\bf Illumination and Pose invariance.}
  Pose and illumination have been a long standing problem in face recognition.
  This figure shows the output distances of FaceNet between pairs of faces of
  the same and a different person in different pose and illumination
  combinations. A distance of $0.0$ means the faces are identical, $4.0$
corresponds to the opposite spectrum, two different identities. You can see
that a threshold of 1.1 would classify every pair correctly.}
\label{fig:poseillu}
\end{figure}

In this paper we present a unified system for face verification (is this the
same person), recognition (who is this person) and clustering (find common
people among these faces).  Our method is based on learning a Euclidean
embedding per image using a deep convolutional network. The network is trained
such that the squared L2 distances in the embedding space directly correspond to
face similarity: faces of the same person have small distances and faces of
distinct people have large distances.

Once this embedding has been produced, then the aforementioned tasks become
straight-forward: face verification simply involves thresholding the distance
between the two embeddings; recognition becomes a k-NN classification problem;
and clustering can be achieved using off-the-shelf techniques such as k-means
or agglomerative clustering.

Previous face recognition approaches based on deep networks use a
classification layer~\cite{deepid2plus,deepface} trained over a set of known
face identities and then take an intermediate bottleneck layer as a
representation used to generalize recognition beyond the set of identities used
in training.  The downsides of this approach are its indirectness and its
inefficiency: one has to hope that the bottleneck representation generalizes
well to new faces; and by using a bottleneck layer the representation size per
face is usually very large (1000s of dimensions). Some recent
work~\cite{deepid2plus} has reduced this dimensionality using PCA, but this is
a linear transformation that can be easily learnt in one layer of the network.

In contrast to these approaches, FaceNet directly trains its output to be a
compact \mbox{128-D} embedding using a triplet-based loss function based on
LMNN~\cite{Weinberger06distancemetric}.  Our triplets consist of two matching
face thumbnails and a non-matching face thumbnail and the loss aims to separate
the positive pair from the negative by a distance margin. The thumbnails are
tight crops of the face area, no 2D or 3D alignment, other than scale and
translation is performed.

Choosing which triplets to use turns
out to be very important for achieving good performance and, inspired by
curriculum learning~\cite{Bengio2009Curriculum}, we present a novel online
negative exemplar mining strategy which ensures consistently increasing
difficulty of triplets as the network trains.  To improve
clustering accuracy, we also explore hard-positive mining techniques which
encourage spherical clusters for the embeddings of a single person.

As an illustration of the incredible variability that our method can handle see
Figure~\ref{fig:poseillu}.  Shown are image pairs from PIE~\cite{pie} that
previously were considered to be very difficult for face verification systems.

An overview of the rest of the paper is as follows: in
section~\ref{RelatedWork} we review the literature in this area;
section~\ref{TripletLoss} defines the triplet loss and
section~\ref{TripletSelection} describes our novel triplet selection and
training procedure; in section~\ref{CNN} we describe the model architecture
used.  Finally in section~\ref{Datasets} and~\ref{Experiments} we present some
quantitative results of our embeddings and also qualitatively explore some
clustering results.

\section{Related Work} \label{RelatedWork}

Similarly to other recent works which employ deep
networks~\cite{deepid2plus,deepface}, our approach is a purely data driven
method which learns its representation directly from the pixels of the face.
Rather than using engineered features, we use a large dataset of labelled faces
to attain the appropriate invariances to pose, illumination, and other
variational conditions.

In this paper we explore two different deep network architectures that have
been recently used to great success in the computer vision community.  Both are
deep convolutional networks~\cite{lecun1989backprop,backprop1986}. The first
architecture is based on the Zeiler\&Fergus~\cite{zeilerfergus} model which
consists of multiple interleaved layers of convolutions, non-linear
activations, local response normalizations, and max pooling layers. We
additionally add several $1{\times}1{\times}d$ convolution layers inspired by
the work of~\cite{networkinnetwork}.  The second architecture is based on the
\textit{Inception} model of Szegedy~\etal which was recently used as the
winning approach for ImageNet 2014~\cite{SzegedyILSVRC2014}. These networks use
mixed layers that run several different convolutional and pooling layers in
parallel and concatenate their responses. We have found that these models can
reduce the number of parameters by up to 20 times and have the potential to
reduce the number of FLOPS required for comparable performance.

There is a vast corpus of face verification and recognition works. Reviewing it
is out of the scope of this paper so we will only briefly discuss the most
relevant recent work.

The works of~\cite{deepid2plus,deepface,zhenyao2014} all employ a complex
system of multiple stages, that combines the output of a deep convolutional
network with PCA for dimensionality reduction and an SVM for classification.

Zhenyao~\etal~\cite{zhenyao2014} employ a deep network to ``warp'' faces
into a canonical frontal view and then learn CNN that classifies each face as
belonging to a known identity. For face verification, PCA on the network output
in conjunction with an ensemble of SVMs is used.

Taigman~\etal~\cite{deepface} propose a multi-stage approach that aligns faces
to a general 3D shape model. A multi-class network is trained to perform the
face recognition task on over four thousand identities. The authors also
experimented with a so called Siamese network where they directly optimize the
$L_1$-distance between two face features. Their best performance on LFW
($97.35\%$) stems from an ensemble of three networks using different alignments
and color channels. The predicted distances (non-linear SVM predictions based
on the $\chi^2$ kernel) of those networks are combined using a non-linear SVM.

Sun~\etal~\cite{deepid2,deepid2plus} propose a compact and therefore relatively
cheap to compute network. They use an ensemble of 25 of these network, each
operating on a different face patch. For their final performance on LFW
($99.47\%$~\cite{deepid2plus}) the authors combine 50 responses (regular and
flipped).  Both PCA and a Joint Bayesian model~\cite{chen2012} that effectively
correspond to a linear transform in the embedding space are employed.  Their
method does not require explicit 2D/3D alignment. The networks are trained by
using a combination of classification and verification loss. The verification
loss is similar to the triplet loss we
employ~\cite{Schultz2004,Weinberger06distancemetric}, in that it minimizes the
$L_2$-distance between faces of the same identity and enforces a margin between
the distance of faces of different identities. The main difference is that only
pairs of images are compared, whereas the triplet loss encourages a relative
distance constraint.

A similar loss to the one used here was explored in
Wang~\etal~\cite{Wang2014} for ranking images by semantic and visual
similarity.

\section{Method}

\begin{figure}[t]
\begin{center}
 \includegraphics[trim=1cm 10.5cm 1cm 0cm, width=0.88\linewidth]{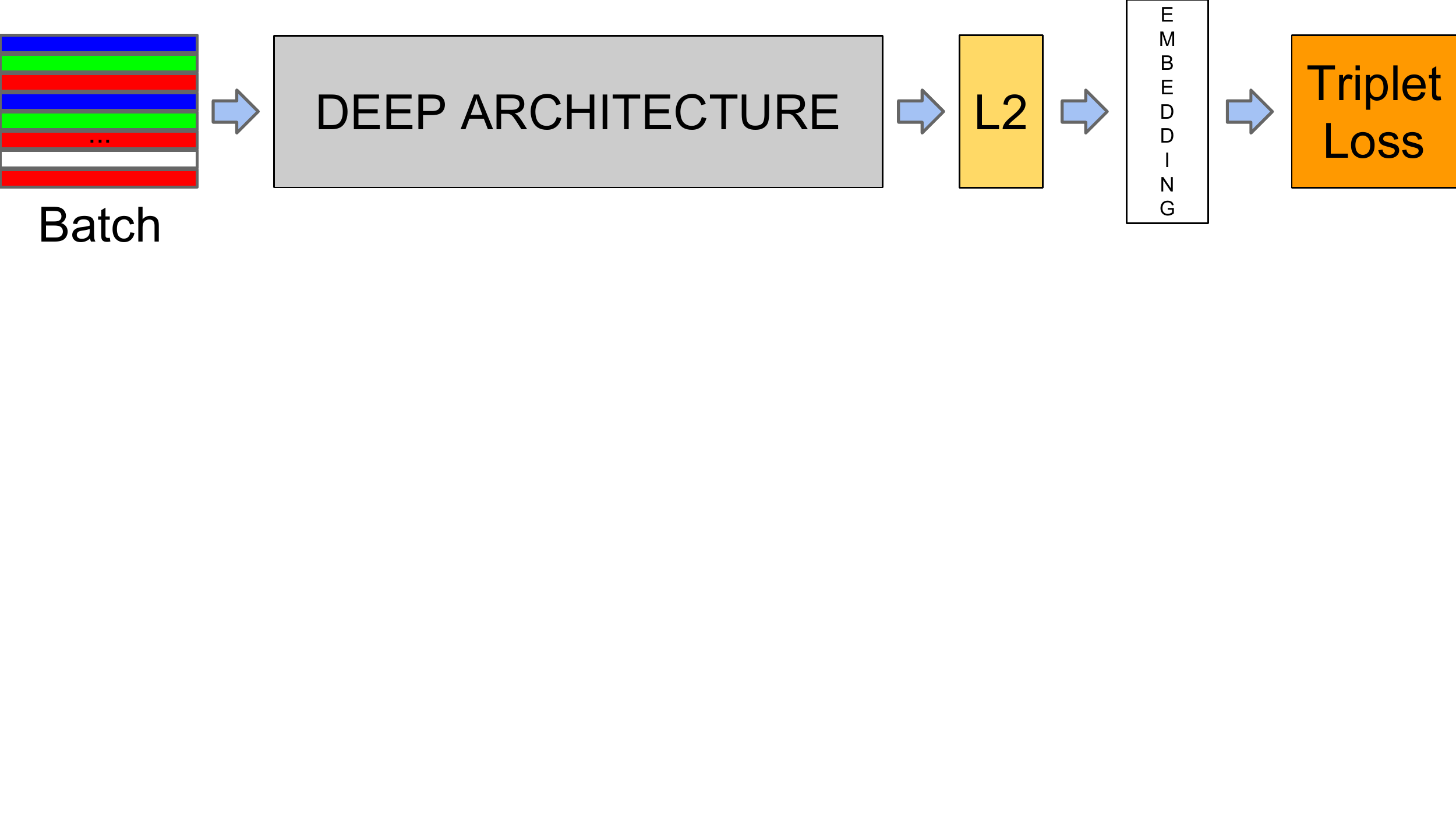}
\end{center}
\caption{{\bf Model structure.} Our network consists of a batch input layer and
  a deep CNN followed by $L_2$ normalization, which results in the face
  embedding. This is followed by the triplet loss during training.}
\label{fig:model}
\end{figure}

FaceNet uses a deep convolutional network. We discuss two different core
architectures: The Zeiler\&Fergus~\cite{zeilerfergus} style networks and the
recent Inception~\cite{SzegedyILSVRC2014} type networks. The details of these
networks are described in section~\ref{architectures}.

Given the model details, and treating it as a black box (see
Figure~\ref{fig:model}), the most important part of our approach lies in the
end-to-end learning of the whole system. To this end we employ the triplet loss
that directly reflects what we want to achieve in face verification,
recognition and clustering. Namely, we strive for an embedding $f(x)$, from an
image $x$ into a feature space $\mathbb{R}^d$, such that the squared distance
between \emph{all} faces, independent of imaging conditions, of the same
identity is small, whereas the squared distance between a pair of face images
from different identities is large.

Although we did not directly compare to other losses, \eg the one using
pairs of positives and negatives, as used in~\cite{deepid2}~Eq.~(2), we believe
that the triplet loss is more suitable for face verification. The motivation is
that the loss from~\cite{deepid2} encourages all faces of one identity to be
projected onto a single point in the embedding space. The triplet loss,
however, tries to enforce a margin between each pair of faces from one person
to all other faces. This allows the faces for one identity to live on a
manifold, while still enforcing the distance and thus discriminability to other
identities.

The following section describes this triplet loss and how it can be learned
efficiently at scale.

\subsection{Triplet Loss}\label{TripletLoss}
\begin{figure}[t]
\begin{center}
 \includegraphics[trim=1cm 10cm 2.5cm 0cm, width=0.9\linewidth]{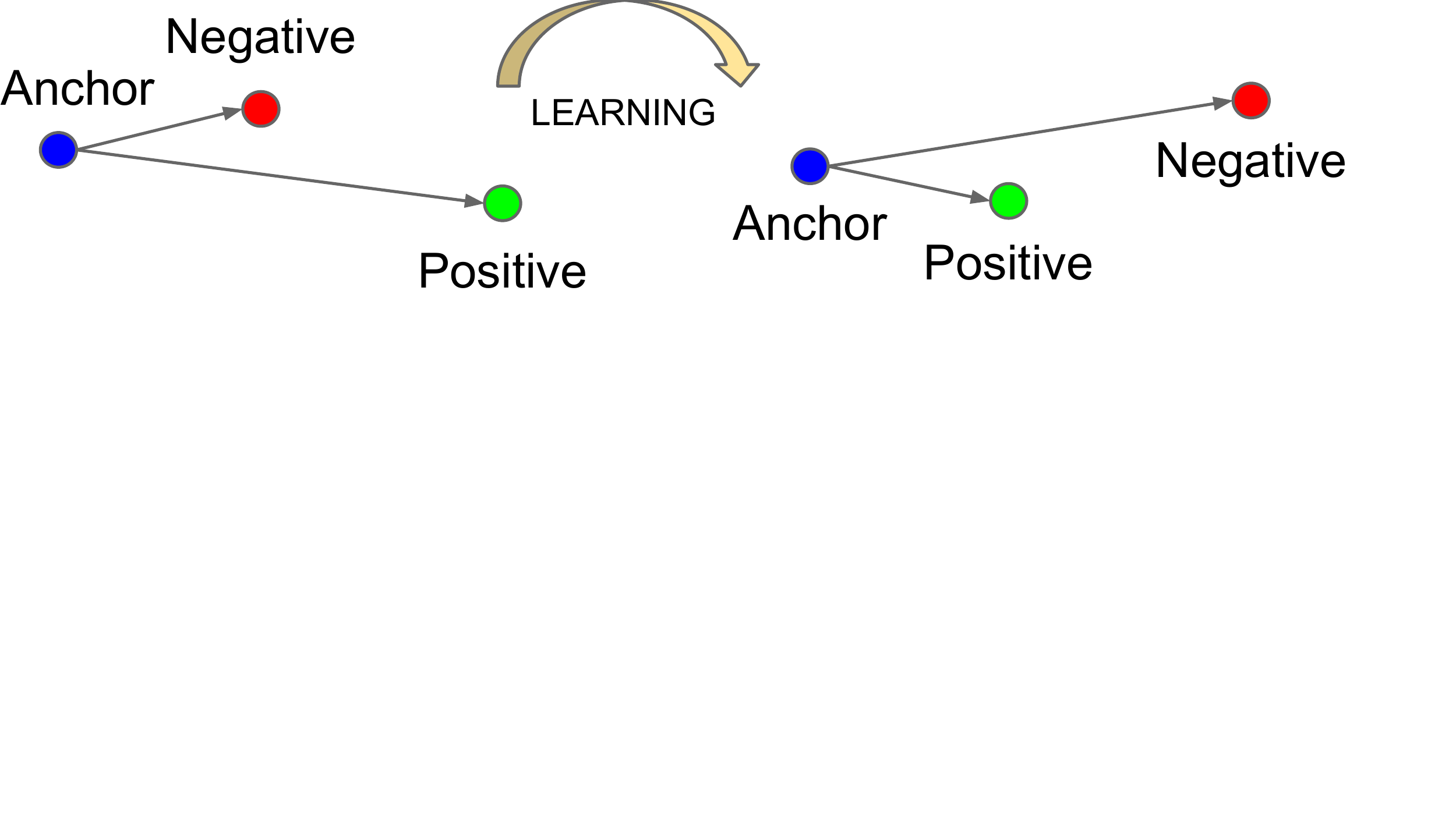}
\end{center}
\caption{The {\bf Triplet Loss} minimizes the distance between an
  \emph{anchor} and a \emph{positive}, both of which have the same identity,
  and maximizes the distance between the \emph{anchor} and a \emph{negative} of
  a different identity.}
\label{fig:triplet_loss_viz}
\end{figure}

The embedding is represented by $f(x)\in\mathbb{R}^d$. It embeds an image $x$
into a \mbox{$d$-dimensional} Euclidean space. Additionally, we constrain this
embedding to live on the \mbox{$d$-dimensional} hypersphere, \ie
$\|f(x)\|_2=1$.  This loss is motivated in~\cite{Weinberger06distancemetric} in
the context of nearest-neighbor classification. Here we want to ensure that an
image $x_i^a$ (\emph{anchor}) of a specific person is closer to all other
images $x_i^p$ (\emph{positive}) of the same person than it is to any image
$x_i^n$ (\emph{negative}) of any other person. This is visualized in
Figure~\ref{fig:triplet_loss_viz}.

Thus we want,
\begin{align}\label{eq:triplet_constraint}
  \|f(x_i^a) - f(x_i^p)\|_2^2 + \alpha &< \|f(x_i^a) - f(x_i^n)\|_2^2\;,\\
  \:\forall\left(f(x_i^a), f(x_i^p), f(x_i^n)\right)\in\mathcal{T}\;.&&
\end{align}
where $\alpha$ is a margin that is enforced between positive and negative
pairs. $\mathcal{T}$ is the set of all possible triplets in the training set
and has cardinality $N$.

The loss that is being minimized is then $L=$
\begin{equation}\label{eq:triplet_loss}
  \sum_i^N\left[\left\|f(x_i^a)-f(x_i^p)\right\|_2^2 -
                  \left\|f(x_i^a)-f(x_i^n)\right\|_2^2+\alpha\right]_+\;.
\end{equation}

Generating all possible triplets would result in many triplets that are easily
satisfied (\ie fulfill the constraint in~Eq.~(\ref{eq:triplet_constraint})).
These triplets would not contribute to the training and result in slower
convergence, as they would still be passed through the network. It is crucial
to select hard triplets, that are active and can therefore contribute to
improving the model. The following section talks about the different approaches
we use for the triplet selection.

\subsection{Triplet Selection} \label{TripletSelection}

In order to ensure fast convergence it is crucial to select triplets that
violate the triplet constraint in~Eq.~(\ref{eq:triplet_constraint}).  This means
that, given $x_i^a$, we want to select an $x_i^p$ (\emph{hard positive}) such
that $\argmax_{x_i^p}\left\|f(x_i^a)-f(x_i^p)\right\|_2^2$ and similarly
$x_i^n$ (\emph{hard negative}) such that $\argmin_{x_i^n}\left\|f(x_i^a)-f(x_i^n)\right\|_2^2$.

It is infeasible to compute the $\argmin$ and $\argmax$ across the whole
training set. Additionally, it might lead to poor training, as mislabelled and
poorly imaged faces would dominate the hard positives and negatives.
There are two obvious choices that avoid this issue:
\begin{itemize}
    \item Generate triplets offline every n~steps, using the most recent
      network checkpoint and computing the $\argmin$ and $\argmax$ on a subset
      of the data.
    \item Generate triplets online.  This can be done by selecting the hard
      positive/negative exemplars from within a mini-batch.
\end{itemize}

Here, we focus on the online generation and use large mini-batches in the order
of a few thousand exemplars and only compute the $\argmin$ and $\argmax$ within
a mini-batch.

To have a meaningful representation of the anchor-positive distances, it needs
to be ensured that a minimal number of exemplars of any one identity is present
in each mini-batch. In our experiments we sample the training data such that
around 40 faces are selected per identity per mini-batch. Additionally, randomly
sampled negative faces are added to each mini-batch.

Instead of picking the hardest positive, we use all anchor-positive pairs in a
mini-batch while still selecting the hard negatives.  We don't have a
side-by-side comparison of hard anchor-positive pairs versus all
anchor-positive pairs within a mini-batch, but we found in practice that the
all anchor-positive method was more stable and converged slightly faster at the
beginning of training.

We also explored the offline generation of triplets in conjunction with the
online generation and it may allow the use of smaller batch sizes, but the
experiments were inconclusive.

Selecting the hardest negatives can in practice lead to bad local minima early on in
training, specifically it can result in a collapsed model (\ie $f(x)=0$). In
order to mitigate this, it helps to select $x_i^n$ such that
\begin{equation}\label{eq:semi_hard}
\left\|f(x_i^a)-f(x_i^p)\right\|_2^2<\left\|f(x_i^a)-f(x_i^n)\right\|_2^2\;.
\end{equation}
We call these negative exemplars \emph{semi-hard}, as they are further away
from the anchor than the positive exemplar, but still hard because the squared
distance is close to the anchor-positive distance. Those negatives lie inside
the margin $\alpha$.

As mentioned before, correct triplet selection is crucial for fast convergence.
On the one hand we would like to use small mini-batches as these tend to
improve convergence during Stochastic Gradient Descent
(SGD)~\cite{Wilson.nn03.batch}.  On the other hand, implementation details make
batches of tens to hundreds of exemplars more efficient. The main constraint
with regards to the batch size, however, is the way we select hard relevant
triplets from within the mini-batches. In most experiments we use a batch size
of around 1,800 exemplars.

\subsection{Deep Convolutional Networks} \label{CNN}
\label{architectures}

\begin{table}
\begin{center}
{\small
\renewcommand\tabcolsep{1.5pt}
\begin{tabular}[H]{|l|c|c|c|c|c|}
\hline
{\bf layer} & {\bf size-in} & {\bf size-out} & {\bf kernel} & {\bf param} & {\bf FLPS} \\
\hline\hline
conv1       & $220{\times}220{\times}3 $ & $110{\times}110{\times}64$ & $7{\times}7{\times}3,2$ & 9K &  115M \\
pool1       & $110{\times}110{\times}64$ & $55{\times}55{\times}64  $ & $3{\times}3{\times}64,2$ & 0 & \\
rnorm1      & $55{\times}55{\times}64  $ & $55{\times}55{\times}64  $ &            & 0 & \\
conv2a      & $55{\times}55{\times}64  $ & $55{\times}55{\times}64  $ & $1{\times}1{\times}64,1$ & 4K &  13M \\
conv2       & $55{\times}55{\times}64  $ & $55{\times}55{\times}192 $ & $3{\times}3{\times}64,1$ & 111K & 335M \\
rnorm2      & $55{\times}55{\times}192 $ & $55{\times}55{\times}192 $ &            & 0 & \\
pool2       & $55{\times}55{\times}192 $ & $28{\times}28{\times}192 $ & $3{\times}3{\times}192,2$ & 0 & \\
conv3a      & $28{\times}28{\times}192 $ & $28{\times}28{\times}192 $ & $1{\times}1{\times}192,1$ & 37K & 29M \\
conv3       & $28{\times}28{\times}192 $ & $28{\times}28{\times}384 $ & $3{\times}3{\times}192,1$ & 664K & 521M \\
pool3       & $28{\times}28{\times}384 $ & $14{\times}14{\times}384 $ & $3{\times}3{\times}384,2$ & 0 & \\
conv4a      & $14{\times}14{\times}384 $ & $14{\times}14{\times}384 $ & $1{\times}1{\times}384,1$ & 148K & 29M \\
conv4       & $14{\times}14{\times}384 $ & $14{\times}14{\times}256 $ & $3{\times}3{\times}384,1$ & 885K & 173M \\
conv5a      & $14{\times}14{\times}256 $ & $14{\times}14{\times}256 $ & $1{\times}1{\times}256,1$ & 66K & 13M \\
conv5       & $14{\times}14{\times}256 $ & $14{\times}14{\times}256 $ & $3{\times}3{\times}256,1$ & 590K & 116M \\
conv6a      & $14{\times}14{\times}256 $ & $14{\times}14{\times}256 $ & $1{\times}1{\times}256,1$ & 66K & 13M \\
conv6       & $14{\times}14{\times}256 $ & $14{\times}14{\times}256 $ & $3{\times}3{\times}256,1$ & 590K & 116M \\
pool4       & $14{\times}14{\times}256 $ & $7{\times}7{\times}256   $ & $3{\times}3{\times}256,2$ & 0 & \\
concat      & $7{\times}7{\times}256   $ & $7{\times}7{\times}256   $ &                      & 0 & \\
fc1         & $7{\times}7{\times}256   $ & $1{\times}32{\times}128  $ & maxout p=2           & 103M & 103M \\
fc2         & $1{\times}32{\times}128  $ & $1{\times}32{\times}128  $ & maxout p=2           & 34M & 34M \\
fc7128      & $1{\times}32{\times}128  $ & $1{\times}1{\times}128   $ &                      & 524K & 0.5M \\
L2 & $1{\times}1{\times}128   $ & $1{\times}1{\times}128   $ &                      & 0 & \\
\hline
\hline
total       &            &            &            & 140M &  1.6B \\
\hline
\end{tabular}
}
\end{center}
\caption{{\bf NN1.} This table show the structure of our
  Zeiler\&Fergus~\cite{zeilerfergus} based model with $1{\times}1$~convolutions
  inspired by~\cite{networkinnetwork}. The input and output sizes are described
  in \mbox{$rows\times cols\times\#filters$}. The kernel is specified as
  \mbox{$rows\times cols, stride$} and the
maxout~\cite{Goodfellow13maxoutnetworks} pooling size as $p=2$.}
\label{tab:nn_zeiler_fergus}
\end{table}

In all our experiments we train the CNN using Stochastic Gradient Descent (SGD)
with standard backprop~\cite{lecun1989backprop,backprop1986} and
AdaGrad~\cite{DuchiAdagrad}. In most experiments we start with a learning rate
of $0.05$ which we lower to finalize the model. The models are initialized from
random, similar to~\cite{SzegedyILSVRC2014}, and trained on a CPU cluster for
1,000 to 2,000 hours. The decrease in the loss (and increase in accuracy) slows
down drastically after 500h of training, but additional training can still
significantly improve performance. The margin $\alpha$ is set to $0.2$.

We used two types of architectures and explore their trade-offs in more detail
in the experimental section. Their practical differences lie in the difference
of parameters and FLOPS. The best model may be different depending on the
application. \Eg a model running in a datacenter can have many parameters and
require a large number of FLOPS, whereas a model running on a mobile phone
needs to have few parameters, so that it can fit into memory. All our models
use rectified linear units as the non-linear activation function.

The first category, shown in Table~\ref{tab:nn_zeiler_fergus}, adds
$1{\times}1{\times}d$ convolutional layers, as suggested
in~\cite{networkinnetwork}, between the standard convolutional layers of the
Zeiler\&Fergus~\cite{zeilerfergus} architecture and results in a model 22
layers deep. It has a total of 140 million parameters and requires around 1.6
billion FLOPS per image.

The second category we use is based on GoogLeNet style Inception
models~\cite{SzegedyILSVRC2014}. These models have $20{\times}$ fewer
parameters (around 6.6M-7.5M) and up to $5{\times}$ fewer FLOPS (between
500M-1.6B). Some of these models are dramatically reduced in size (both depth
and number of filters), so that they can be run on a mobile phone.
One, NNS1, has 26M parameters and only requires 220M FLOPS per image. The
other, NNS2, has 4.3M parameters and 20M FLOPS. Table~\ref{tab:nn2} describes
NN2 our largest network in detail. NN3 is identical in architecture but has a
reduced input size of 160x160. NN4 has an input size of only 96x96, thereby
drastically reducing the CPU requirements (285M FLOPS vs 1.6B for NN2). In
addition to the reduced input size it does not use 5x5 convolutions in the
higher layers as the receptive field is already too small by then. Generally we
found that the 5x5 convolutions can be removed throughout with only a minor
drop in accuracy. Figure~\ref{fig:scatter_plot_flops} compares all our models.

\begin{table*}
\begin{center}
{\small
\begin{tabular}[H]{|l|c|c|c|c|c|c|c|c|c|c|c|}
\hline
{\bf type} & {\bf \stackanchor{output}{size}} & {\bf depth} & {\bf $\#1{\times}1$} & {\bf \stackanchor{$\#3{\times}3$}{reduce}} & $\#3{\times}3$ & {\bf \stackanchor{$\#5{\times}5$}{reduce}} & $\#5{\times}5$ & {\bf \stackanchor{pool}{proj (p)}} & {\bf params} & {\bf FLOPS} \\
\hline\hline
conv1 ($7{\times}7{\times}3,2$) & $112{\times}112{\times}64$ & 1 & & & & & & & 9K & 119M \\
\hline
max pool + norm & $56{\times}56{\times}64$ & 0 & & & & & & m $3{\times}3,2$ & & \\
\hline
inception (2) & $56{\times}56{\times}192$ & 2 & & 64 & 192 & & & & 115K & 360M \\
\hline
norm + max pool & $28{\times}28{\times}192$ & 0 & & & & & & m $3{\times}3,2$ & & \\
\hline
inception (3a) & $28{\times}28{\times}256$ & 2 & 64 & 96 & 128 & 16 & 32 & m, 32p & 164K & 128M \\
\hline
inception (3b) & $28{\times}28{\times}320$ & 2 & 64 & 96 & 128 & 32 & 64 & $L_2$, 64p & 228K & 179M \\
\hline
inception (3c) & $14{\times}14{\times}640$ & 2 & 0 & 128 & 256,2 & 32 & 64,2 & m $3{\times}3$,2 & 398K & 108M \\
\hline
inception (4a) & $14{\times}14{\times}640$ & 2 & 256 & 96 & 192 & 32 & 64 & $L_2$, 128p & 545K & 107M \\
\hline
inception (4b) & $14{\times}14{\times}640$ & 2 & 224 & 112 & 224 & 32 & 64 & $L_2$, 128p & 595K & 117M \\
\hline
inception (4c) & $14{\times}14{\times}640$ & 2 & 192 & 128 & 256 & 32 & 64 & $L_2$, 128p & 654K & 128M \\
\hline
inception (4d) & $14{\times}14{\times}640$ & 2 & 160 & 144 & 288 & 32 & 64 & $L_2$, 128p & 722K & 142M \\
\hline
inception (4e) & $7{\times}7{\times}1024$ & 2 & 0 & 160 & 256,2 & 64 & 128,2 & m $3{\times}3$,2 & 717K & 56M \\
\hline
inception (5a) & $7{\times}7{\times}1024$ & 2 & 384 & 192 & 384 & 48 & 128 & $L_2$, 128p & 1.6M & 78M \\
\hline
inception (5b) & $7{\times}7{\times}1024$ & 2 & 384 & 192 & 384 & 48 & 128 & m, 128p & 1.6M & 78M \\
\hline
avg pool & $1{\times}1{\times}1024$ & 0 & & & & & & & & \\
\hline
fully conn & $1{\times}1{\times}128$ & 1 & & & & & & & 131K & 0.1M \\
\hline
L2 normalization & $1{\times}1{\times}128$ & 0 & & & & & & & & \\
\hline
\hline
total & & & & & & & & & 7.5M & 1.6B \\
\hline
\end{tabular}
}
\end{center}
\caption{{\bf NN2.} Details of the NN2 Inception incarnation. This model is
  almost identical to the one described in ~\cite{SzegedyILSVRC2014}. The two
  major differences are the use of $L_2$ pooling instead of max pooling (m),
  where specified. \Ie instead of taking the spatial max the $L_2$ norm is
  computed. The pooling is always $3{\times}3$ (aside from the final average
  pooling) and in parallel to the convolutional modules inside each Inception
  module. If there is a dimensionality reduction after the pooling it is
denoted with p. $1{\times}1$, $3{\times}3$, and $5{\times}5$ pooling are then
concatenated to get the final output.}
\label{tab:nn2}
\end{table*}


\section{Datasets and Evaluation}
\label{Datasets}

We evaluate our method on four datasets and with the exception of Labelled
Faces in the Wild and YouTube Faces we evaluate our method on the face
verification task. \Ie given a pair of two face images a squared $L_2$ distance
threshold $D(x_i,x_j)$ is used to determine the classification of \emph{same}
and \emph{different}. All faces pairs $(i,j)$ of the same identity are denoted
with $\mathcal{P_\text{same}}$, whereas all pairs of different identities are
denoted with $\mathcal{P_\text{diff}}$.

We define the set of all \emph{true accepts} as
\begin{equation}
  \ta(d)=\{(i,j)\in\mathcal{P_\text{same}},\text{with}\,D(x_i,x_j)\leq d\}\;.
\end{equation}
These are the face pairs $(i,j)$ that were correctly classified as \emph{same}
at threshold $d$. Similarly
\begin{equation}
  \fa(d)=\{(i,j)\in\mathcal{P_\text{diff}},\text{with}\,D(x_i,x_j)\leq d\}
\end{equation}
is the set of all pairs that was incorrectly classified as \emph{same}
(\emph{false accept}).

The validation rate $\val(d)$ and the false accept rate $\far(d)$ for a given
face distance $d$ are then defined as
\begin{equation}
  \val(d)=\frac{\left|\ta(d)\right|}{\left|\mathcal{P_\text{same}}\right|}\;,\quad
  \far(d)=\frac{\left|\fa(d)\right|}{\left|\mathcal{P_\text{diff}}\right|}\;.
\end{equation}

\subsection{Hold-out Test Set}
\label{hold_out_test}
We keep a hold out set of around one million images, that has the same
distribution as our training set, but disjoint identities. For evaluation we
split it into five disjoint sets of $200\text{k}$ images each. The $\far$ and
$\val$ rate are then computed on $100\text{k}\times100\text{k}$ image pairs.
Standard error is reported across the five splits.

\subsection{Personal Photos}
\label{personal_photos_test}
This is a test set with similar distribution to our training set, but has been
manually verified to have very clean labels. It consists of three personal
photo collections with a total of around $12\text{k}$ images. We compute the
$\far$ and $\val$ rate across all $\text{12k}$ squared pairs of images.

\subsection{Academic Datasets}

Labeled Faces in the Wild (LFW) is the de-facto academic test set for face
verification~\cite{lfw}. We follow the standard protocol for
\emph{unrestricted, labeled outside data} and report the mean classification
accuracy as well as the standard error of the mean.

Youtube Faces DB~\cite{ytf} is a new dataset that has gained popularity in
the face recognition community~\cite{deepface,deepid2plus}. The setup is
similar to LFW, but instead of verifying pairs of images, pairs of videos are
used.

\section{Experiments} \label{Experiments}

If not mentioned otherwise we use between 100M-200M training face thumbnails
consisting of about 8M different identities. A face detector is run on each
image and a tight bounding box around each face is generated. These face
thumbnails are resized to the input size of the respective network. Input sizes
range from 96x96 pixels to 224x224 pixels in our experiments.

\subsection{Computation Accuracy Trade-off}

\begin{figure}[t]
\begin{center}
 \includegraphics[trim=1cm .5cm 1cm .5cm, width=0.9\linewidth]{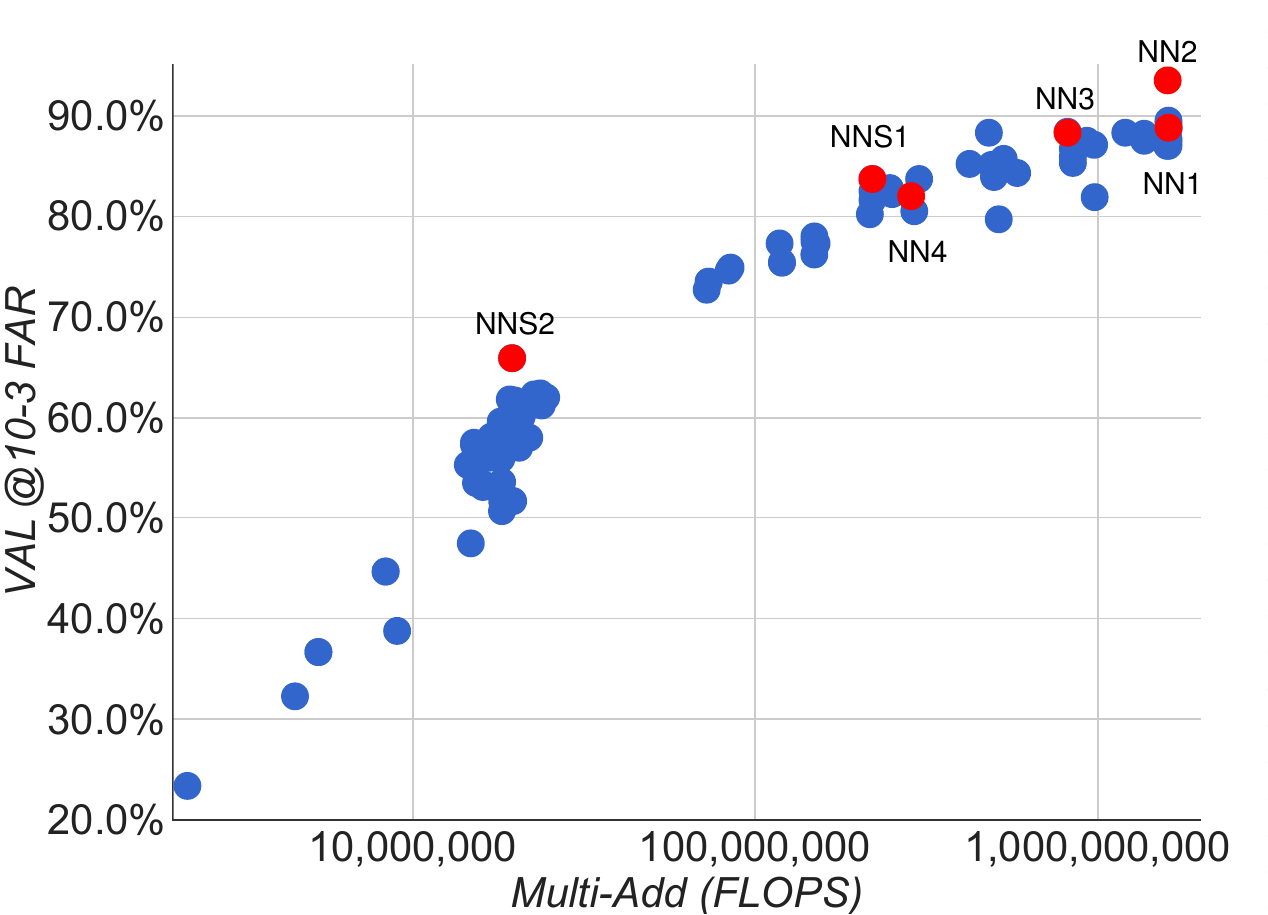}
\end{center}
\caption{{\bf FLOPS \vs Accuracy trade-off.} Shown is the trade-off between
  FLOPS and accuracy for a wide range of different model sizes and
  architectures. Highlighted are the four models that we focus on in our
  experiments.}
\label{fig:scatter_plot_flops}
\end{figure}

Before diving into the details of more specific experiments we will discuss the
trade-off of accuracy versus number of FLOPS that a particular model requires.
Figure~\ref{fig:scatter_plot_flops} shows the FLOPS on the x-axis and the
accuracy at 0.001 false accept rate ($\far$) on our user labelled test-data set
from section~\ref{personal_photos_test}. It is interesting to see the strong
correlation between the computation a model requires and the accuracy it
achieves. The figure highlights the five models (NN1, NN2, NN3, NNS1, NNS2)
that we discuss in more detail in our experiments.

We also looked into the accuracy trade-off with regards to the number of model
parameters. However, the picture is not as clear in that case. For example, the
Inception based model NN2 achieves a comparable performance to NN1, but only
has a 20th of the parameters. The number of FLOPS is comparable, though.
Obviously at some point the performance is expected to decrease, if the number
of parameters is reduced further. Other
model architectures may allow further reductions without loss of
accuracy, just like Inception~\cite{SzegedyILSVRC2014} did in this case.

\subsection{Effect of CNN Model}

\begin{figure}[t]
\begin{center}
 \includegraphics[trim=1cm .5cm 1cm 0cm, width=0.9\linewidth]{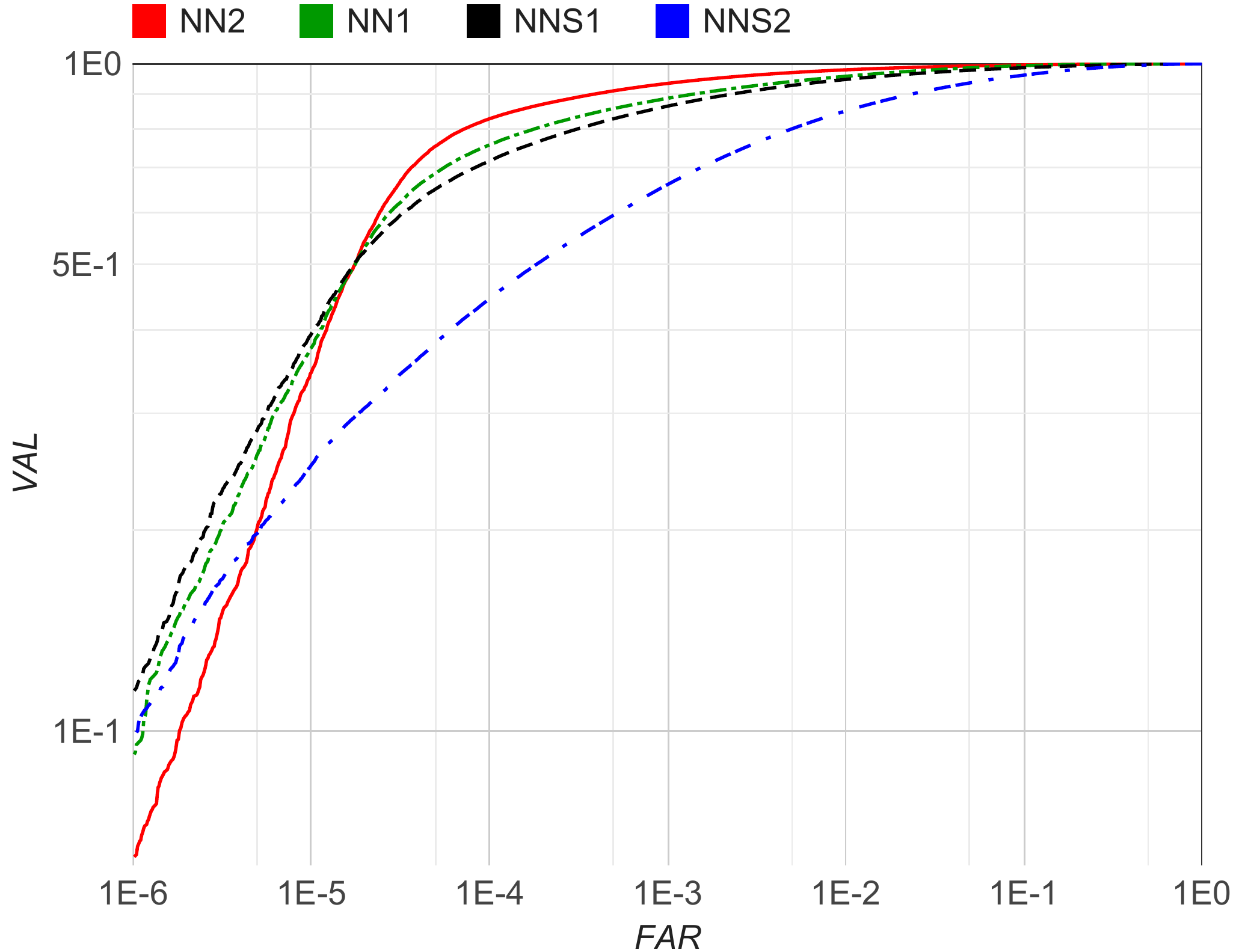}
\end{center}
\caption{{\bf Network Architectures.} This plot shows the complete ROC for the
  four different models on our personal photos test set from
  section~\ref{personal_photos_test}. The sharp drop at $10\text{\sc{e}-4}$
$\far$ can be explained by noise in the groundtruth labels. The models in order of performance are:
\mbox{{\bf NN2:} $224{\times}224$ input} Inception based model;
\mbox{{\bf NN1:} Zeiler\&Fergus} based network with $1{\times}1$ convolutions;
\mbox{{\bf NNS1:} small} Inception style model with only 220M FLOPS;
\mbox{{\bf NNS2:} tiny} Inception model with only 20M FLOPS.}
\label{fig:network_architectures}
\end{figure}

\begin{table}
\begin{center}
\begin{tabular}{|l|c|}
\hline
architecture & $\val$ \\
\hline\hline
NN1 (Zeiler\&Fergus $220{\times}220$) & $87.9\%\pm1.9$ \\ 
NN2 (Inception $224{\times}224$) & $89.4\%\pm1.6$ \\ 
NN3 (Inception $160{\times}160$) & $88.3\%\pm1.7$ \\ 
NN4 (Inception $96{\times}96$)& $82.0\%\pm2.3$ \\ 
NNS1 (mini Inception $165{\times}165$)& $82.4\%\pm2.4$ \\ 
NNS2 (tiny Inception $140{\times}116$)& $51.9\%\pm2.9$ \\ 
\hline
\end{tabular}
\end{center}
\caption{{\bf Network Architectures.} This table compares the performance of
our model architectures on the hold out test set (see
section~\ref{hold_out_test}). Reported is the mean validation rate $\val$ at
$10\text{\sc{e}-3}$ false accept rate. Also shown is the standard error of the
mean across the five test splits.}
\label{tab:network_architectures}
\end{table}

We now discuss the performance of our four selected models in more detail. On
the one hand we have our traditional Zeiler\&Fergus based architecture with
$1{\times}1$ convolutions~\cite{zeilerfergus,networkinnetwork} (see
Table~\ref{tab:nn_zeiler_fergus}). On the other hand we have
Inception~\cite{SzegedyILSVRC2014} based models that dramatically reduce the
model size.  Overall, in the final performance the top models of both
architectures perform comparably. However, some of our Inception based models,
such as NN3, still achieve good performance while significantly reducing both
the FLOPS and the model size.

The detailed evaluation on our personal photos
test set is shown in Figure~\ref{fig:network_architectures}. While the largest
model achieves a dramatic improvement in accuracy compared to the tiny
NNS2, the latter can be run 30ms / image on a mobile phone and is still accurate
enough to be used in face clustering. The sharp drop in the ROC for
$\far<10^{-4}$ indicates noisy labels in the test data groundtruth. At extremely
low false accept rates a single mislabeled image can have a significant impact
on the curve.

\subsection{Sensitivity to Image Quality}

\begin{table}
\begin{center}
\begin{tabular}{|c|c|}
\hline
jpeg q & val-rate \\
\hline\hline
10 & 67.3\% \\
20 & 81.4\% \\
30 & 83.9\% \\
50 & 85.5\% \\
70 & 86.1\% \\
90 & 86.5\% \\
\hline
\end{tabular}
\hspace{.5cm}
\begin{tabular}{|r|c|}
\hline
\#pixels & val-rate \\
\hline\hline
1,600 & 37.8\% \\
6,400 & 79.5\% \\
14,400 & 84.5\% \\
25,600 & 85.7\% \\
65,536 & 86.4\% \\
\hline
\end{tabular}
\end{center}
\caption{{\bf Image Quality.} The table on the left shows the effect on the
  validation rate at $10\text{\sc{e}-3}$ precision with varying JPEG quality.
  The one on the right shows how the image size in pixels effects the
  validation rate at $10\text{\sc{e}-3}$ precision. This experiment was done
  with NN1 on the first split of our test hold-out dataset.}
\label{tab:image_quality}
\end{table}

Table~\ref{tab:image_quality} shows the robustness of our model across a wide
range of image sizes. The network is surprisingly robust with respect to JPEG
compression and performs very well down to a JPEG quality of 20. The performance
drop is very small for face thumbnails down to a size of 120x120 pixels and
even at 80x80 pixels it shows acceptable performance. This is notable, because
the network was trained on 220x220 input images. Training with lower resolution
faces could improve this range further.

\subsection{Embedding Dimensionality}

\begin{table}
\begin{center}
\begin{tabular}{|r|c|}
\hline
\#dims & $\val$ \\
\hline\hline
64 & $86.8\%\pm1.7$ \\
128 & $87.9\%\pm1.9$ \\
256 & $87.7\%\pm1.9$ \\
512 & $85.6\%\pm2.0$ \\
\hline
\end{tabular}
\end{center}
\caption{{\bf Embedding Dimensionality.} This Table compares the effect of the
  embedding dimensionality of our model NN1 on our hold-out set from
  section~\ref{hold_out_test}. In addition to the $\val$ at
  $10\text{\sc{e}-3}$ we also show the standard error of the mean computed
  across five splits.}
\label{tab:embedding_dims}
\end{table}
We explored various embedding dimensionalities and selected 128 for all
experiments other than the comparison reported in
Table~\ref{tab:embedding_dims}. One would expect the larger embeddings to
perform at least as good as the smaller ones, however, it is possible that they
require more training to achieve the same accuracy. That said, the differences
in the performance reported in Table~\ref{tab:embedding_dims} are statistically
insignificant.

It should be noted, that during training a 128 dimensional float vector is
used, but it can be quantized to \mbox{128-bytes} without loss of accuracy. Thus each
face is compactly represented by a 128 dimensional byte vector, which is ideal
for large scale clustering and recognition. Smaller embeddings are possible at
a minor loss of accuracy and could be employed on mobile devices.

\subsection{Amount of Training Data}

\begin{table}
\begin{center}
\begin{tabular}{|c|c|}
\hline
\#training images & $\val$ \\
\hline\hline
2,600,000 & 76.3\% \\
26,000,000 & 85.1\% \\
52,000,000 & 85.1\% \\
260,000,000 & 86.2\% \\
\hline
\end{tabular}
\end{center}
\caption{{\bf Training Data Size.} This table compares the performance after
  700h of training for a smaller model with 96x96 pixel inputs. The model
  architecture is similar to NN2, but without the 5x5 convolutions in the
  Inception modules.}
\label{tab:training_data}
\end{table}

Table~\ref{tab:training_data} shows the impact of large amounts of training
data. Due to time constraints this evaluation was run on a smaller model;
the effect may be even larger on larger models. It is clear that using tens
of millions of exemplars results in a clear boost of accuracy on our personal
photo test set from section~\ref{personal_photos_test}. Compared to only
millions of images the relative reduction in error is 60\%. Using another order
of magnitude more images (hundreds of millions) still gives a small boost, but
the improvement tapers off.

\subsection{Performance on LFW}
\label{lfw_experiments}

\begin{figure}[t]
\begin{center}
\renewcommand\tabcolsep{1pt}
{\bf False accept}
\begin{tabular}{cc||cc||cc}
  \includegraphics[width=.15\linewidth]{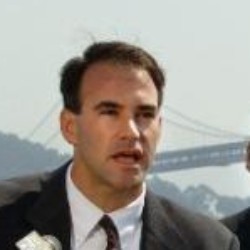} & \includegraphics[width=.15\linewidth]{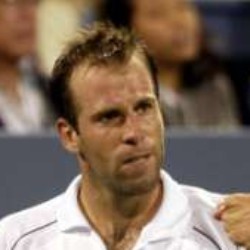} &
  \includegraphics[width=.15\linewidth]{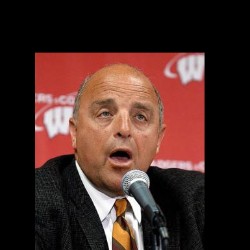} & \includegraphics[width=.15\linewidth]{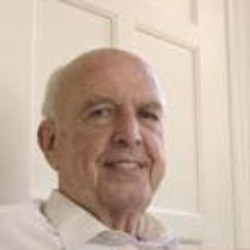} &
  \includegraphics[width=.15\linewidth]{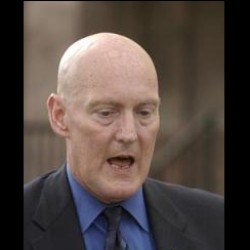} & \includegraphics[width=.15\linewidth]{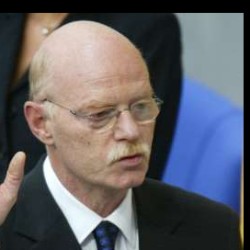} \\
  \includegraphics[width=.15\linewidth]{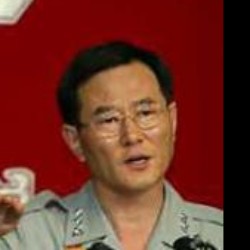} & \includegraphics[width=.15\linewidth]{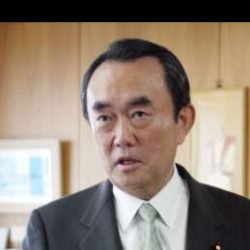} &
  \includegraphics[width=.15\linewidth]{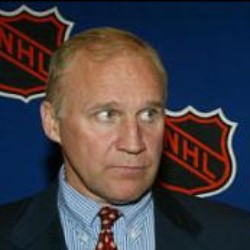} & \includegraphics[width=.15\linewidth]{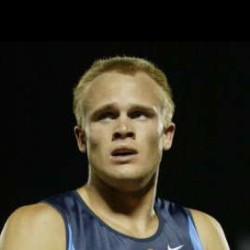} &
  \includegraphics[width=.15\linewidth]{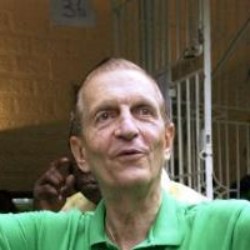} & \includegraphics[width=.15\linewidth]{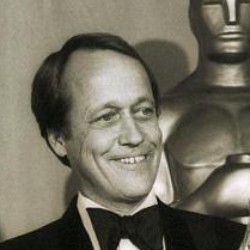} \\
  \includegraphics[width=.15\linewidth]{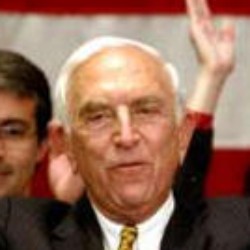} & \includegraphics[width=.15\linewidth]{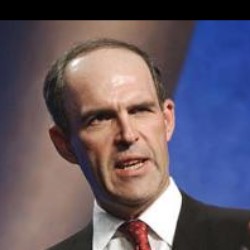} &
  \includegraphics[width=.15\linewidth]{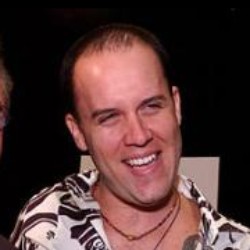} & \includegraphics[width=.15\linewidth]{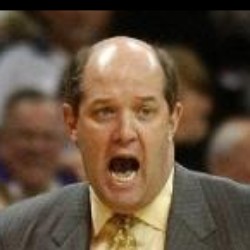} &
  \includegraphics[width=.15\linewidth]{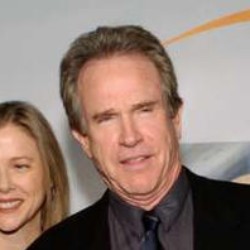} & \includegraphics[width=.15\linewidth]{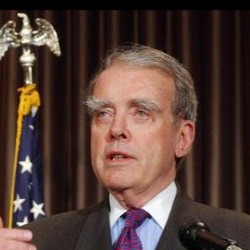} \\
\end{tabular}
{\bf False reject}
\begin{tabular}{cc||cc||cc}
  \includegraphics[width=.15\linewidth]{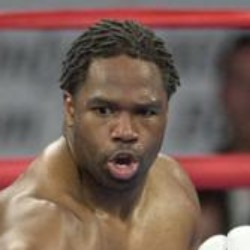} & \includegraphics[width=.15\linewidth]{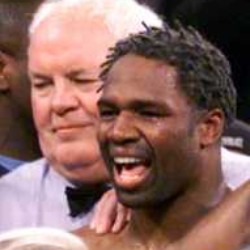} &
  \includegraphics[width=.15\linewidth]{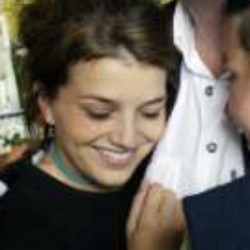} & \includegraphics[width=.15\linewidth]{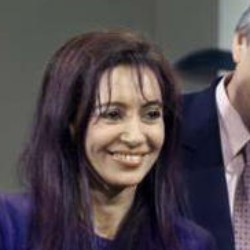} &
  \includegraphics[width=.15\linewidth]{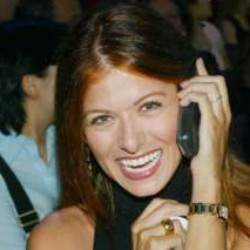} & \includegraphics[width=.15\linewidth]{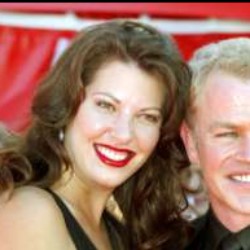} \\
  \includegraphics[width=.15\linewidth]{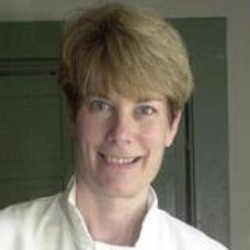} & \includegraphics[width=.15\linewidth]{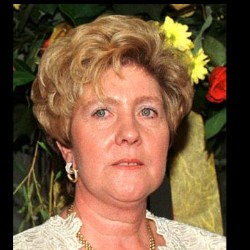} &
  \includegraphics[width=.15\linewidth]{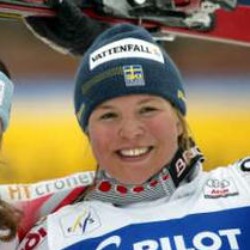} & \includegraphics[width=.15\linewidth]{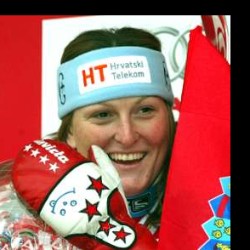} &
  \includegraphics[width=.15\linewidth]{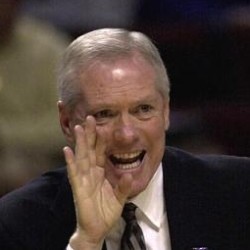} & \includegraphics[width=.15\linewidth]{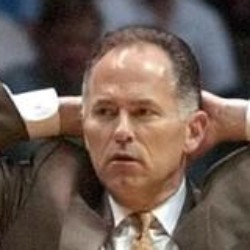} \\
  \includegraphics[width=.15\linewidth]{figures/Jim_OBrien_0001.jpg} & \includegraphics[width=.15\linewidth]{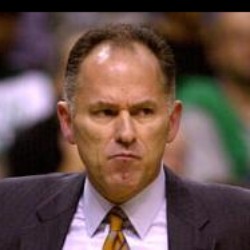} &
  \includegraphics[width=.15\linewidth]{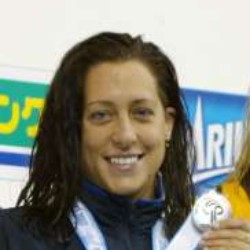} & \includegraphics[width=.15\linewidth]{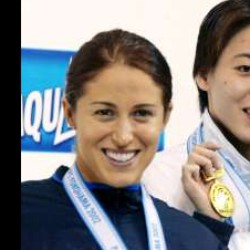} &
  \includegraphics[width=.15\linewidth]{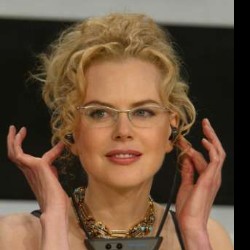} & \includegraphics[width=.15\linewidth]{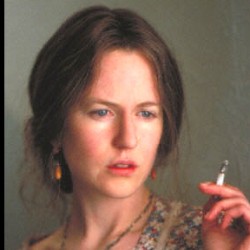} \\
  \includegraphics[width=.15\linewidth]{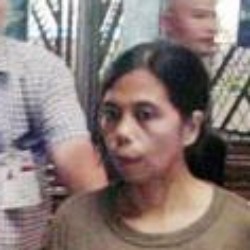} & \includegraphics[width=.15\linewidth]{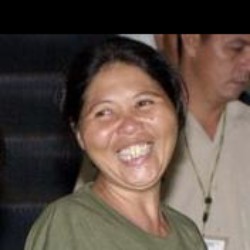} &
  \includegraphics[width=.15\linewidth]{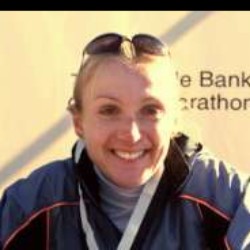} & \includegraphics[width=.15\linewidth]{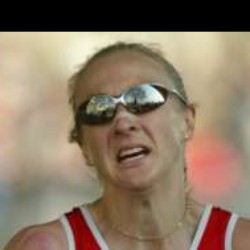} &
  \includegraphics[width=.15\linewidth]{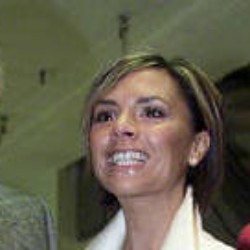} & \includegraphics[width=.15\linewidth]{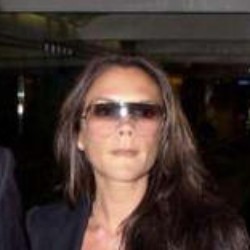} \\
  \includegraphics[width=.15\linewidth]{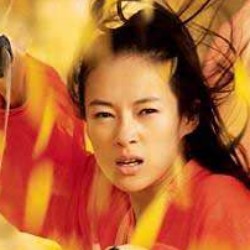} & \includegraphics[width=.15\linewidth]{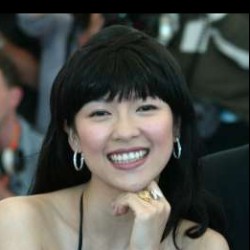} \\
\end{tabular}
\end{center}
\caption{{\bf LFW errors.} This shows all pairs of images that were incorrectly
classified on LFW. Only eight of the 13 false rejects shown here are actual errors the
other five are mislabeled in LFW.}
\label{fig:lfw_errors}
\end{figure}

We evaluate our model on LFW using the standard protocol for
\emph{unrestricted, labeled outside data}. Nine training splits are used to
select the $L_2$-distance threshold. Classification
(\emph{same} or \emph{different}) is then performed on the tenth test split.
The selected optimal threshold is $1.242$ for all test splits except split
eighth ($1.256$).

Our model is evaluated in two modes:
\begin{enumerate}
    \item Fixed center crop of the LFW provided thumbnail.\vspace{-.5em}
    \item A proprietary face detector (similar to
      Picasa~\cite{joint_cascade_face_detection}) is run on the provided LFW
      thumbnails. If it fails to align the face (this happens for two images),
      the LFW alignment is used.
\end{enumerate}

Figure~\ref{fig:lfw_errors} gives an overview of \emph{all} failure cases. It
shows false accepts on the top as well as false rejects at the bottom.  We
achieve a classification accuracy of \textbf{98.87\%}$\pm0.15$ when using the
fixed center crop described in (1) and the record breaking
\textbf{99.63\%}$\pm$0.09 standard error of the mean when using the extra face
alignment (2). This reduces the error reported for DeepFace in~\cite{deepface}
by more than a factor of 7 and the previous state-of-the-art reported for
DeepId2+ in~\cite{deepid2plus} by 30\%. This is the performance of model NN1,
but even the much smaller NN3 achieves performance that is not statistically
significantly different.

\subsection{Performance on Youtube Faces DB}
\label{ytf_experiments}
We use the average similarity of all pairs of the first one hundred frames
that our face detector detects in each video. This gives us a classification
accuracy of \textbf{95.12\%}$\pm0.39$.  Using the first one thousand frames
  results in 95.18\%.
  Compared to~\cite{deepface}~91.4\% who also evaluate one hundred frames per
  video we reduce the error rate by almost half. DeepId2+~\cite{deepid2plus}
  achieved 93.2\% and our method reduces this error by 30\%, comparable to our
  improvement on LFW.

\subsection{Face Clustering}

\begin{figure}[t]
\begin{center}
 \includegraphics[trim=0cm 0cm 0cm 0cm, width=0.9\linewidth]{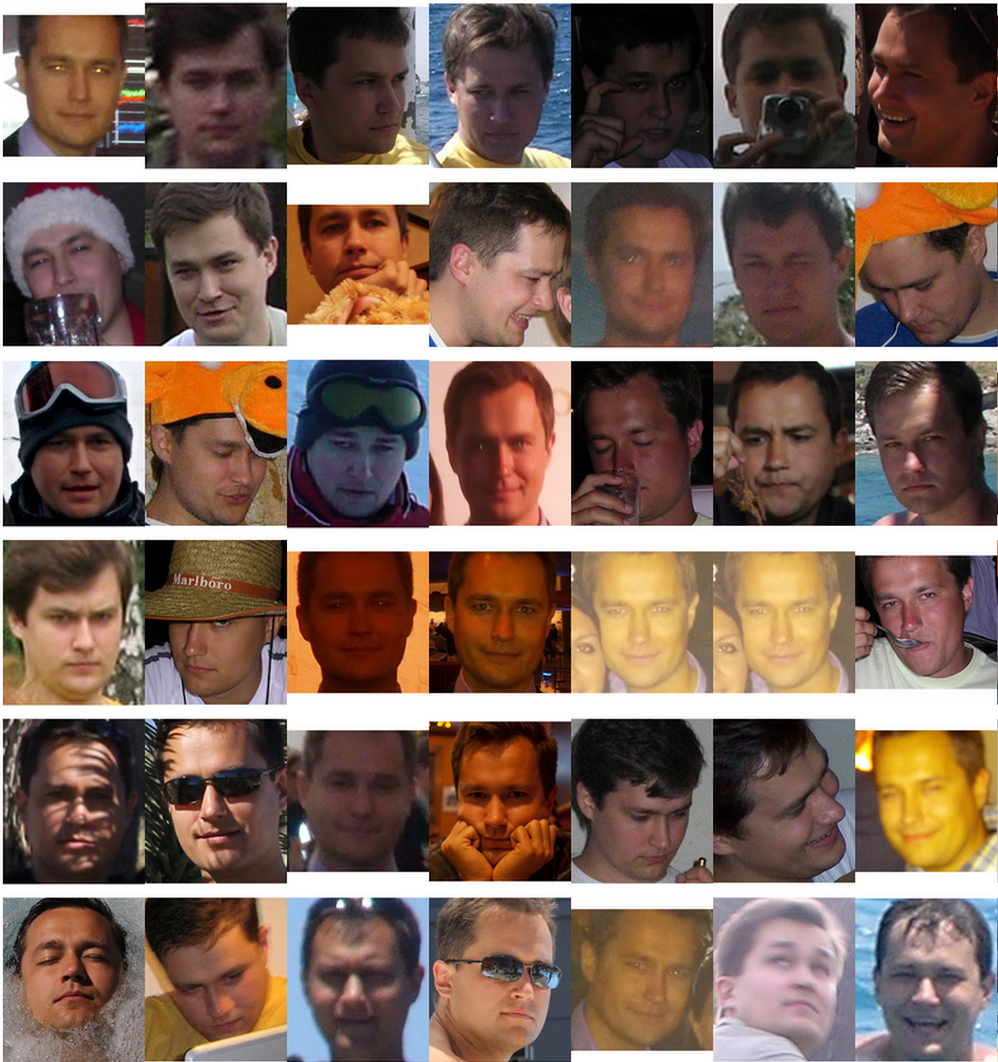}
 \includegraphics[trim=0cm 1cm 0cm 0cm, width=0.9\linewidth]{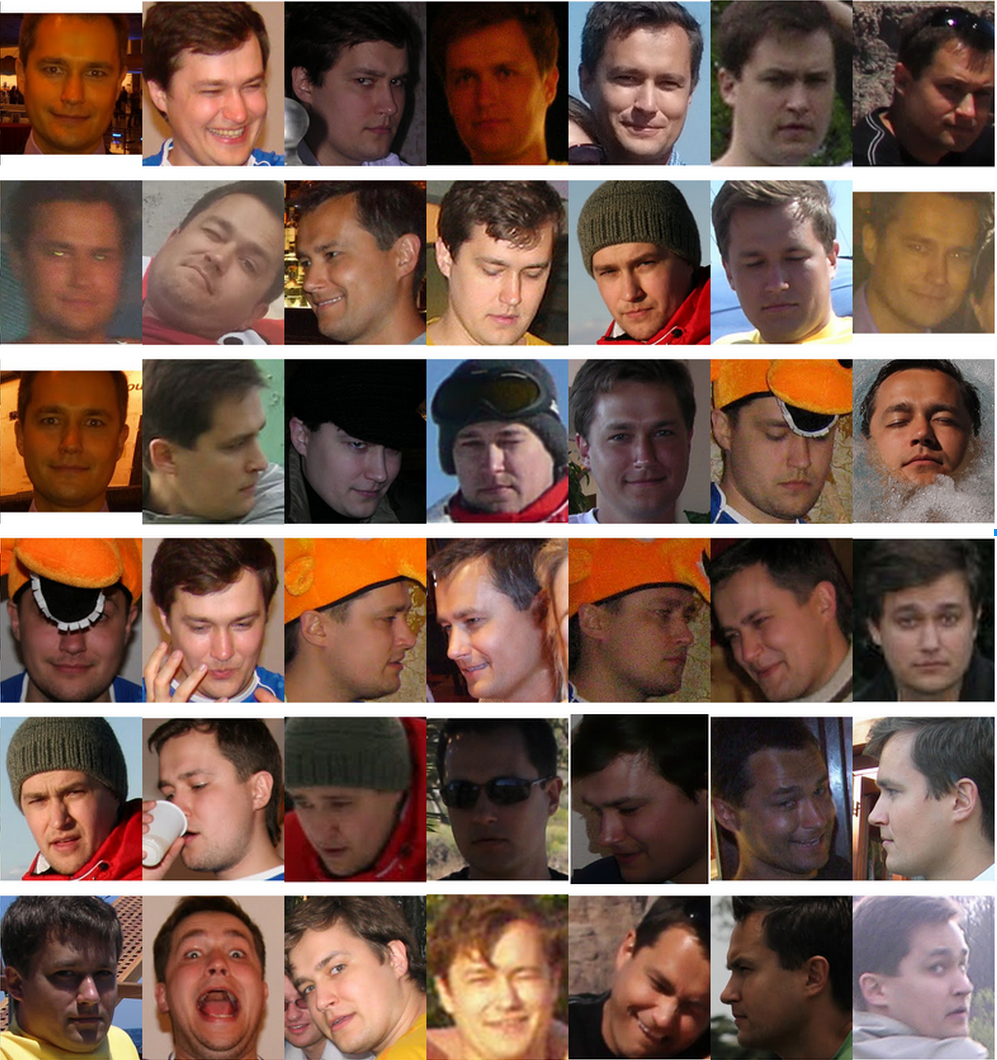}
\end{center}
\caption{{\bf Face Clustering.} Shown is an exemplar cluster for one user. All
  these images in the users personal photo collection were clustered
  together.}
\label{fig:face_clustering}
\end{figure}

Our compact embedding lends itself to be used in order to cluster a users
personal photos into groups of people with the same identity. The constraints
in assignment imposed by clustering faces, compared to the pure verification
task, lead to truly amazing results.  Figure~\ref{fig:face_clustering} shows
one cluster in a users personal photo collection, generated using agglomerative
clustering.  It is a clear showcase of the incredible invariance to occlusion,
lighting, pose and even age.

\section{Summary}

We provide a method to directly learn an embedding into an Euclidean space for
face verification. This sets it apart from other
methods~\cite{deepid2plus,deepface} who use the CNN bottleneck layer, or
require additional post-processing such as concatenation of multiple models and
PCA, as well as SVM classification.  Our end-to-end training both simplifies
the setup and shows that directly optimizing a loss relevant to the task at
hand improves performance.

Another strength of our model is that it only requires minimal alignment (tight
crop around the face area). \cite{deepface}, for example, performs a complex 3D
alignment. We also experimented with a similarity transform alignment and
notice that this can actually improve performance slightly. It is not clear if
it is worth the extra complexity.

Future work will focus on better understanding of the error cases, further
improving the model, and also reducing model size and reducing CPU
requirements. We will also look into ways of improving the currently extremely
long training times, \eg variations of our curriculum learning with smaller
batch sizes and offline as well as online positive and negative mining.

\section{Appendix: Harmonic Embedding}
In this section we introduce the concept of \emph{harmonic} embeddings. By this
we denote a set of embeddings that are generated by different models v1 and v2
but are compatible in the sense that they can be compared to each other.

\begin{figure}[t]
\begin{center}
 \includegraphics[trim=1.5cm 1cm 2cm 0cm, width=0.9\linewidth]{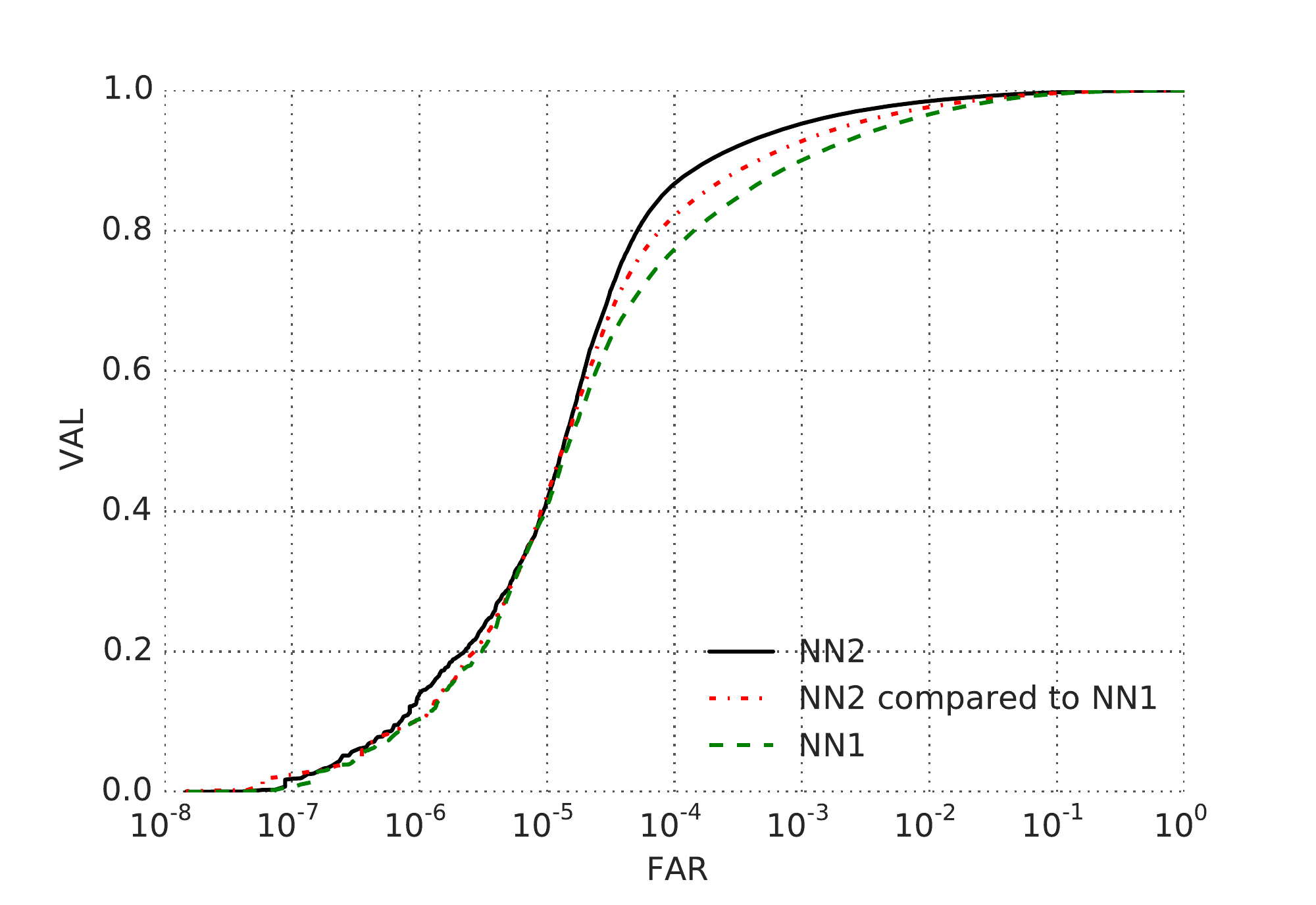}
\end{center}
\caption{{\bf Harmonic Embedding Compatibility.} These ROCs show the
compatibility of the \emph{harmonic} embeddings of NN2 to the embeddings of
NN1. NN2 is an improved model that performs much better than NN1. When
comparing embeddings generated by NN1 to the \emph{harmonic} ones generated by
NN2 we can see the compatibility between the two. In fact, the mixed mode
performance is still better than NN1 by itself.}
\label{fig:harmonic_rocs}
\end{figure}

This compatibility greatly simplifies upgrade paths. \Eg~in an scenario where
embedding v1 was computed across a large set of images and a new embedding
model v2 is being rolled out, this compatibility ensures a smooth transition
without the need to worry about version incompatibilities.
Figure~\ref{fig:harmonic_rocs} shows results on our 3G dataset. It can be seen
that the improved model NN2 significantly outperforms NN1, while the comparison
of NN2 embeddings to NN1 embeddings performs at an intermediate level.

\subsection{Harmonic Triplet Loss}
\begin{figure}[t]
\begin{center}
 \includegraphics[trim=0cm .5cm 0cm 0cm, width=0.7\linewidth]{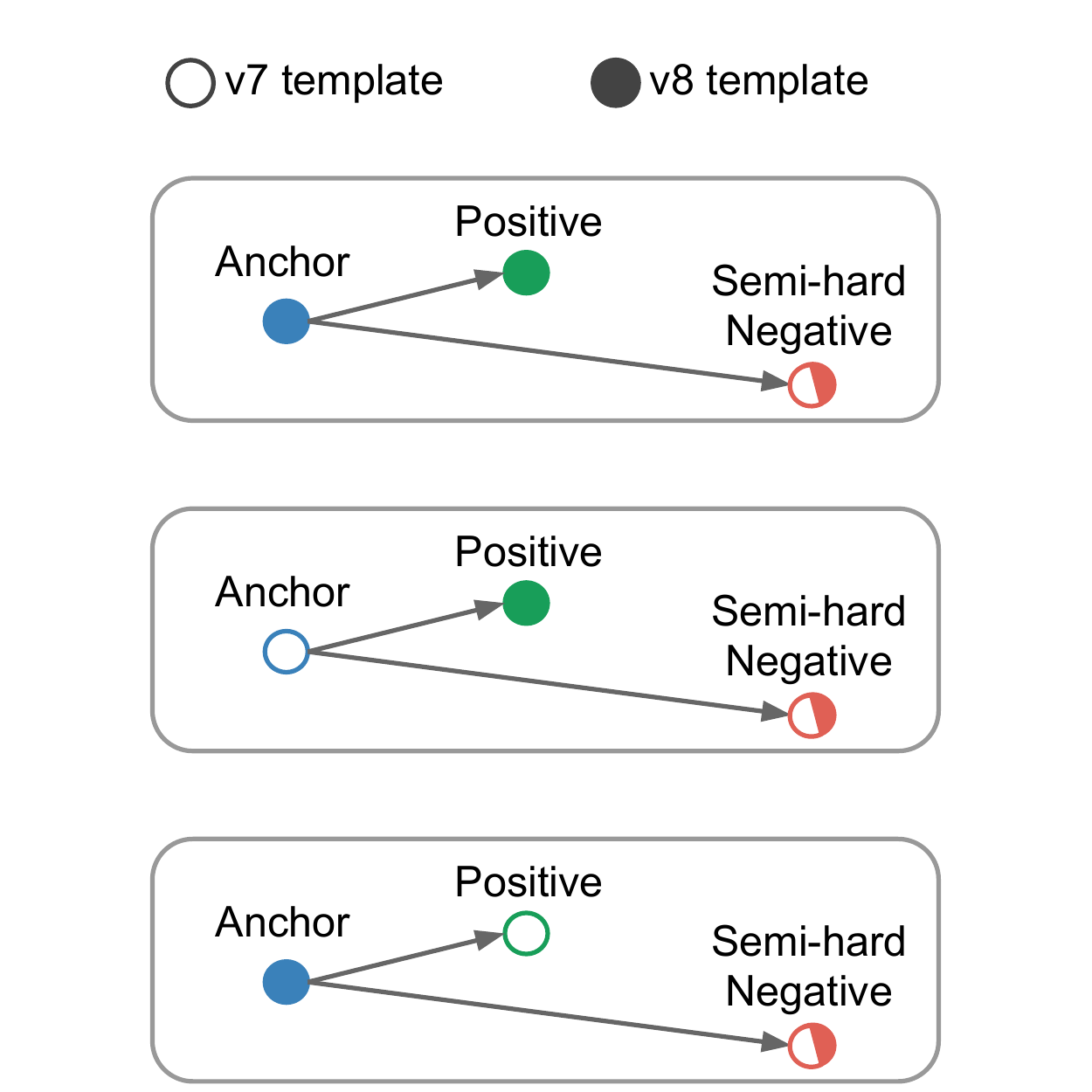}
\end{center}
\caption{{\bf Learning the Harmonic Embedding.} In order to learn a
\emph{harmonic} embedding, we generate triplets that mix the v1 embeddings with
the v2 embeddings that are being trained. The semi-hard negatives are selected
from the whole set of both v1 and v2 embeddings.}
\label{fig:harmonic_embedding_training}
\end{figure}

In order to learn the \emph{harmonic} embedding we mix embeddings of v1
together with the embeddings v2, that are being learned. This is done inside
the triplet loss and results in additionally generated triplets that encourage
the compatibility between the different embedding versions.
Figure~\ref{fig:harmonic_embedding_training} visualizes the different
combinations of triplets that contribute to the triplet loss.

We initialized the v2 embedding from an independently trained NN2 and retrained
the last layer (embedding layer) from random initialization with the
compatibility encouraging triplet loss. First only the last layer is retrained,
then we continue training the whole v2 network with the harmonic loss.

\begin{figure}[t]
\begin{center}
 \includegraphics[trim=.5cm 7cm 0cm 0cm, width=0.9\linewidth]{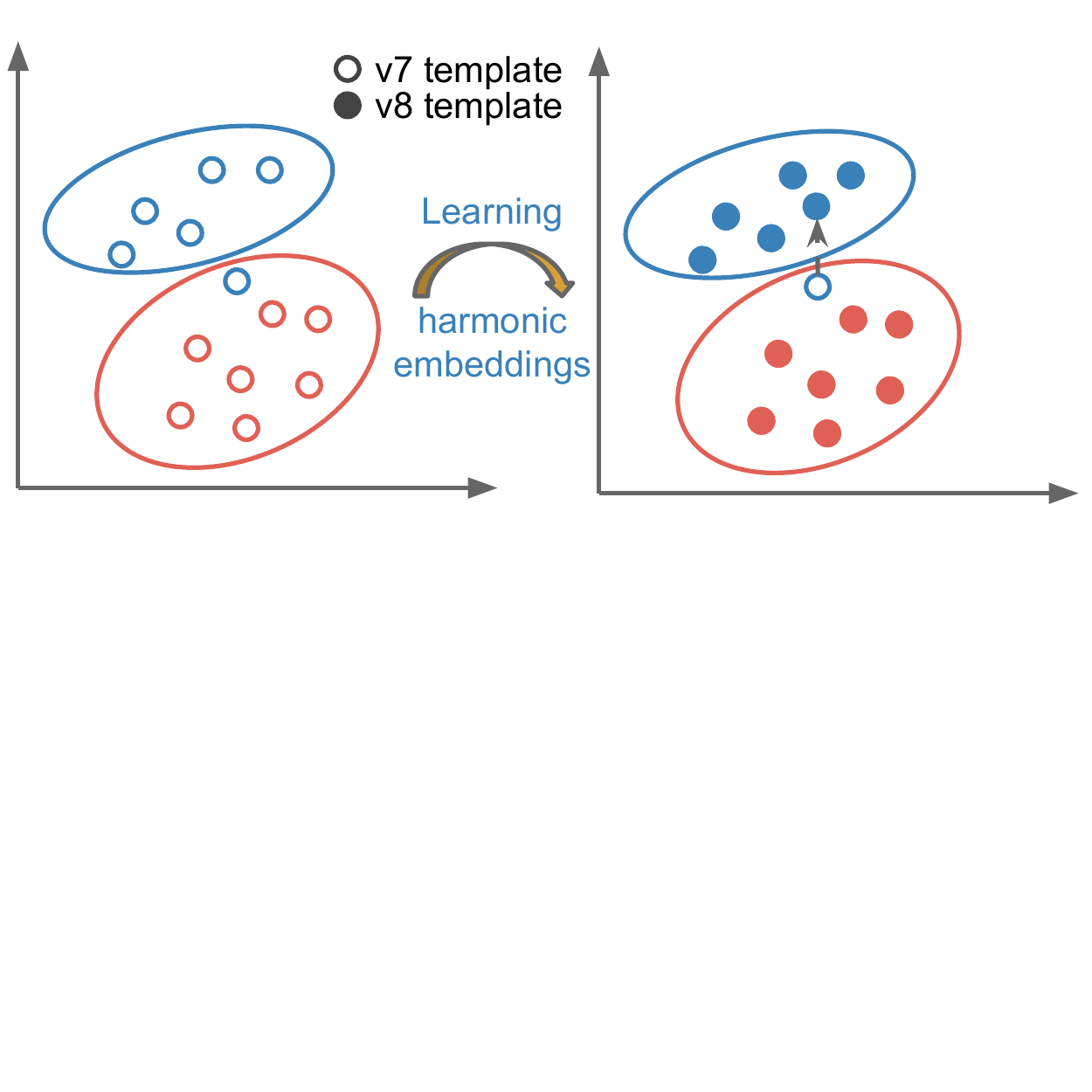}
\end{center}
\caption{{\bf Harmonic Embedding Space.} This visualisation sketches a possible
interpretation of how \emph{harmonic} embeddings are able to improve
verification accuracy while maintaining compatibility to less accurate
embeddings. In this scenario there is one misclassified face, whose embedding
is perturbed to the ``correct'' location in v2.}
\label{fig:harmonic_embedding_space}
\end{figure}

Figure~\ref{fig:harmonic_embedding_space} shows a possible interpretation of
how this compatibility may work in practice. The vast majority of v2 embeddings
may be embedded near the corresponding v1 embedding, however, incorrectly
placed v1 embeddings can be perturbed slightly such that their new location in
embedding space improves verification accuracy.

\subsection{Summary}
These are very interesting findings and it is somewhat surprising that it works
so well. Future work can explore how far this idea can be extended. Presumably
there is a limit as to how much the v2 embedding can improve over v1, while
still being compatible. Additionally it would be interesting to train small
networks that can run on a mobile phone and are compatible to a larger server
side model.

\section*{Acknowledgments}
We would like to thank Johannes Steffens for his discussions and great insights
on face recognition and Christian Szegedy for providing new network
architectures like~\cite{SzegedyILSVRC2014} and discussing network design
choices.  Also we are indebted to the DistBelief~\cite{dean2012large} team for
their support especially to Rajat Monga for help in setting up efficient
training schemes.

Also our work would not have been possible without the support of Chuck
Rosenberg, Hartwig Adam, and Simon Han.

{\small
\bibliographystyle{ieee}
\bibliography{facenet}

\begin{thebibliography}{10}\itemsep=-1pt

\bibitem{Bengio2009Curriculum}
Y.~Bengio, J.~Louradour, R.~Collobert, and J.~Weston.
\newblock Curriculum learning.
\newblock In {\em Proc. of ICML}, New York, NY, USA, 2009.

\bibitem{chen2012}
D.~Chen, X.~Cao, L.~Wang, F.~Wen, and J.~Sun.
\newblock Bayesian face revisited: A joint formulation.
\newblock In {\em Proc. ECCV}, 2012.

\bibitem{joint_cascade_face_detection}
D.~Chen, S.~Ren, Y.~Wei, X.~Cao, and J.~Sun.
\newblock Joint cascade face detection and alignment.
\newblock In {\em Proc. ECCV}, 2014.

\bibitem{dean2012large}
J.~Dean, G.~Corrado, R.~Monga, K.~Chen, M.~Devin, M.~Mao, M.~Ranzato,
  A.~Senior, P.~Tucker, K.~Yang, Q.~V. Le, and A.~Y. Ng.
\newblock Large scale distributed deep networks.
\newblock In P.~Bartlett, F.~Pereira, C.~Burges, L.~Bottou, and K.~Weinberger,
  editors, {\em NIPS}, pages 1232--1240. 2012.

\bibitem{DuchiAdagrad}
J.~Duchi, E.~Hazan, and Y.~Singer.
\newblock Adaptive subgradient methods for online learning and stochastic
  optimization.
\newblock {\em J. Mach. Learn. Res.}, 12:2121--2159, July 2011.

\bibitem{Goodfellow13maxoutnetworks}
I.~J. Goodfellow, D.~Warde-farley, M.~Mirza, A.~Courville, and Y.~Bengio.
\newblock Maxout networks.
\newblock In {\em In ICML}, 2013.

\bibitem{lfw}
G.~B. Huang, M.~Ramesh, T.~Berg, and E.~Learned-Miller.
\newblock Labeled faces in the wild: A database for studying face recognition
  in unconstrained environments.
\newblock Technical Report 07-49, University of Massachusetts, Amherst, October
  2007.

\bibitem{lecun1989backprop}
Y.~LeCun, B.~Boser, J.~S. Denker, D.~Henderson, R.~E. Howard, W.~Hubbard, and
  L.~D. Jackel.
\newblock Backpropagation applied to handwritten zip code recognition.
\newblock {\em Neural Computation}, 1(4):541--551, Dec. 1989.

\bibitem{networkinnetwork}
M.~Lin, Q.~Chen, and S.~Yan.
\newblock Network in network.
\newblock {\em CoRR}, abs/1312.4400, 2013.

\bibitem{gaussianface}
C.~Lu and X.~Tang.
\newblock Surpassing human-level face verification performance on {LFW} with
  gaussianface.
\newblock {\em CoRR}, abs/1404.3840, 2014.

\bibitem{backprop1986}
D.~E. Rumelhart, G.~E. Hinton, and R.~J. Williams.
\newblock Learning representations by back-propagating errors.
\newblock {\em Nature}, 1986.

\bibitem{Schultz2004}
M.~Schultz and T.~Joachims.
\newblock Learning a distance metric from relative comparisons.
\newblock In S.~Thrun, L.~Saul, and B.~Sch\"{o}lkopf, editors, {\em NIPS},
  pages 41--48. MIT Press, 2004.

\bibitem{pie}
T.~Sim, S.~Baker, and M.~Bsat.
\newblock The {CMU} pose, illumination, and expression {(PIE)} database.
\newblock In {\em In Proc. FG}, 2002.

\bibitem{deepid2}
Y.~Sun, X.~Wang, and X.~Tang.
\newblock Deep learning face representation by joint
  identification-verification.
\newblock {\em CoRR}, abs/1406.4773, 2014.

\bibitem{deepid2plus}
Y.~Sun, X.~Wang, and X.~Tang.
\newblock Deeply learned face representations are sparse, selective, and
  robust.
\newblock {\em CoRR}, abs/1412.1265, 2014.

\bibitem{SzegedyILSVRC2014}
C.~Szegedy, W.~Liu, Y.~Jia, P.~Sermanet, S.~Reed, D.~Anguelov, D.~Erhan,
  V.~Vanhoucke, and A.~Rabinovich.
\newblock Going deeper with convolutions.
\newblock {\em CoRR}, abs/1409.4842, 2014.

\bibitem{deepface}
Y.~Taigman, M.~Yang, M.~Ranzato, and L.~Wolf.
\newblock Deepface: Closing the gap to human-level performance in face
  verification.
\newblock In {\em IEEE Conf. on CVPR}, 2014.

\bibitem{Wang2014}
J.~Wang, Y.~Song, T.~Leung, C.~Rosenberg, J.~Wang, J.~Philbin, B.~Chen, and
  Y.~Wu.
\newblock Learning fine-grained image similarity with deep ranking.
\newblock {\em CoRR}, abs/1404.4661, 2014.

\bibitem{Weinberger06distancemetric}
K.~Q. Weinberger, J.~Blitzer, and L.~K. Saul.
\newblock Distance metric learning for large margin nearest neighbor
  classification.
\newblock In {\em NIPS}. MIT Press, 2006.

\bibitem{Wilson.nn03.batch}
D.~R. Wilson and T.~R. Martinez.
\newblock The general inefficiency of batch training for gradient descent
  learning.
\newblock {\em Neural Networks}, 16(10):1429--1451, 2003.

\bibitem{ytf}
L.~Wolf, T.~Hassner, and I.~Maoz.
\newblock Face recognition in unconstrained videos with matched background
  similarity.
\newblock In {\em IEEE Conf. on CVPR}, 2011.

\bibitem{zeilerfergus}
M.~D. Zeiler and R.~Fergus.
\newblock Visualizing and understanding convolutional networks.
\newblock {\em CoRR}, abs/1311.2901, 2013.

\bibitem{zhenyao2014}
Z.~Zhu, P.~Luo, X.~Wang, and X.~Tang.
\newblock Recover canonical-view faces in the wild with deep neural networks.
\newblock {\em CoRR}, abs/1404.3543, 2014.

\end{thebibliography}
}

\end{document}